\documentclass[10pt,twocolumn,letterpaper]{article}

\usepackage[pagenumbers]{cvpr} 

\usepackage{multirow}
\usepackage{makecell}
\usepackage{colortbl}
\newcommand{\cmark}{\ding{52}}%
\newcommand{\xmark}{\ding{56}}%
\usepackage{setspace}
\usepackage{calligra}
\usepackage{algpseudocode}
\usepackage{arydshln}
\usepackage{tabu}
\usepackage{pifont}

\definecolor{cvprblue}{rgb}{0.21,0.49,0.74}
\usepackage[pagebackref,breaklinks,colorlinks,allcolors=cvprblue]{hyperref}

\def\paperID{13451} 
\def\confName{CVPR}
\def\confYear{2025}

\title{AlignGPT: Multi-modal Large Language Models with Adaptive \\Alignment Capability}

\author{Fei Zhao\thanks{Equal contributions.}, Taotian Pang\footnotemark[1], Chunhui Li, Zhen Wu\thanks{Corresponding author.}, Junjie Guo, Shangyu Xing, Xinyu Dai\\
National Key Laboratory for Novel Software Technology, Nanjing University\\
{\tt\small \{zhaof,pangtt,lich,guojj,xsy\}@smail.nju.edu.cn,\{wuz,daixinyu\}@nju.edu.cn}\\
\url{https://aligngpt-vl.github.io}
}

\begin{document}
\maketitle
\begin{abstract}
Multimodal Large Language Models (MLLMs) are widely regarded as crucial in the exploration of Artificial General Intelligence (AGI). The core of MLLMs lies in their capability to achieve cross-modal alignment. To attain this goal, current MLLMs typically follow a two-phase training paradigm: the pre-training phase and the instruction-tuning phase. Despite their success, there are shortcomings in the modeling of alignment capabilities within these models. Firstly, during the pre-training phase, the model usually assumes that all image-text pairs are uniformly aligned, but in fact the degree of alignment between different image-text pairs is inconsistent. Secondly, the instructions currently used for finetuning incorporate a variety of tasks and different tasks usually require different levels of alignment capabilities, but previous MLLMs overlook these differentiated alignment needs. To tackle these issues, we propose a new multimodal large language model AlignGPT. In the pre-training stage, instead of treating all image-text pairs equally, we divide them into different groups according to the degrees of alignment of them. Then, the model is trained to learn the representations of different alignment levels. In the instruction-tuning phase, we adaptively combine these representations of alignment levels to meet the dynamic alignment needs of different tasks. Extensive experimental results show that our model achieves competitive performance on 12 benchmarks.
\end{abstract}
\section{Introduction}
\label{sec:intro}

Multimodal Large Language Models (MLLMs) are considered a crucial step towards achieving Artificial General Intelligence (AGI)~\cite{DBLP:journals/corr/abs-2303-08774,DBLP:journals/corr/abs-2312-11805,DBLP:conf/icml/DriessXSLCIWTVY23,DBLP:journals/corr/abs-2309-16058}. The uniqueness of these models lies in their ability to integrate and understand various types of information, especially text and image data. In the pursuit of AGI, this cross-modal understanding and processing capability is essential, as it mimics how humans interact with the world and comprehend complex information through different senses, such as vision and language. The development of multimodal large language models not only advances the field of artificial intelligence but also provides machines with a way to process and understand information that is closer to human cognition.

\begin{figure}[t]
\centering
    \includegraphics[width=0.47\textwidth]{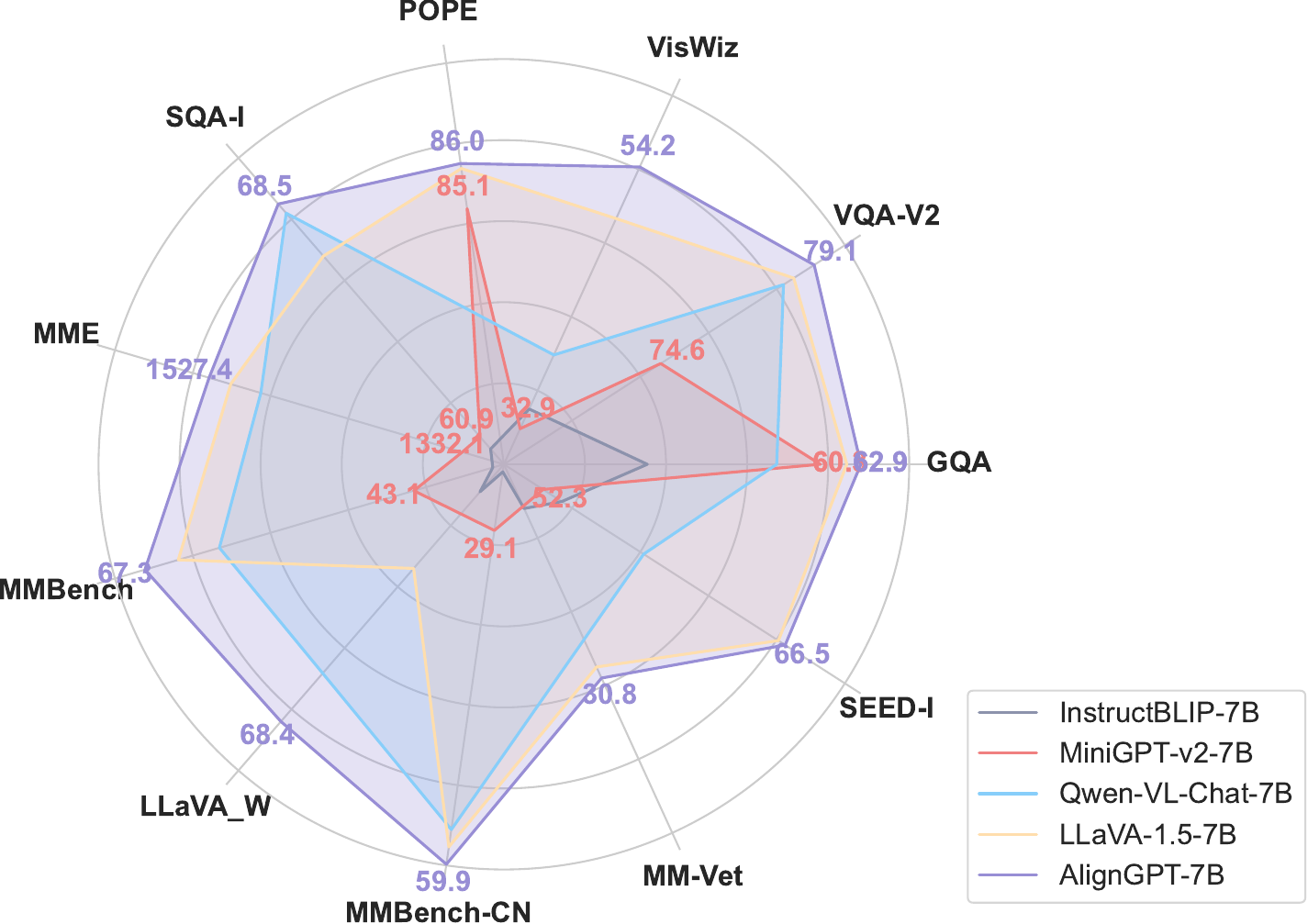}
  \caption{AlignGPT achieves competitive performances on a broad range of vision-language tasks compared with other generalist models. To facilitate observation, we only show the performance of MiniGPT-v2 and AlignGPT.}
  \label{fig:AlignGPT_example}
\end{figure}

\begin{figure*}[t]
\centering
    \includegraphics[width=0.98\textwidth]{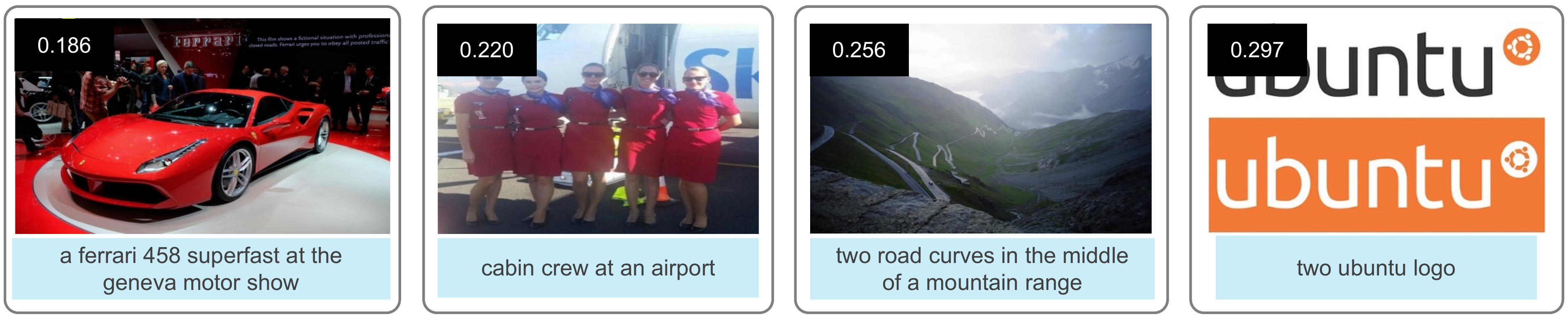}
  \caption{Examples of image-text pairs in the pre-training dataset, where the numbers in each image represent the CLIP similarity.}
  \label{fig:CLIP_Similarity}
\end{figure*}

Currently, MLLMs typically adhere to a unified training paradigm, which is divided into two key phases: the pre-training phase and the instruction-tuning phase~\cite{DBLP:conf/nips/LiuLWL23a,DBLP:journals/corr/abs-2304-10592,DBLP:journals/corr/abs-2304-14178,DBLP:journals/corr/abs-2308-12966,DBLP:journals/corr/abs-2310-09478,DBLP:journals/corr/abs-2311-03079,DBLP:conf/aaai/HuXLLCT24,DBLP:journals/corr/abs-2309-05519,DBLP:journals/corr/abs-2309-15112}. The pre-training phase concentrates on aligning images with text, aiming to train the model to understand the relation between image contents and the respective textual descriptions of them. This alignment imbues the model with cross-modal comprehension abilities. The instruction-tuning phase further enhances its adaptability to specific tasks. This includes enabling the model to complete particular visual-language tasks based on given instructions, such as generating textual descriptions from images, answering questions related to images, or even performing complex reasoning based on both text and images. This training paradigm equips multimodal pre-trained models with not only fundamental cross-modal understanding but also the flexibility to adapt to diverse task demands.

Although current MLLMs have made great progress, the modeling of alignment capabilities of them is still insufficient for the following reasons: 

\begin{itemize}
\item \textbf{The degree of alignment is inconsistent between different image-text pairs:} During the pre-training phase, the model typically operates on a key assumption that all image-text pairs are consistently aligned. However, the degree of alignment in image-text pairs is not always uniform: in some image-text pairs, the text may describe the whole image (as shown in the rightmost of \cref{fig:CLIP_Similarity}) while in others the text only describes a part of the image (as shown in the left three image-text pairs in \cref{fig:CLIP_Similarity}) . If these differences are not differentiated during the pre-training phase, it could lead to a misunderstanding of the image-text alignment relationships in the learning process.

\item \textbf{The different tasks require different levels of alignment capabilities:} The instructions currently used for finetuning cover a variety of tasks. Some of them, like image captioning~\cite{DBLP:conf/icml/XuBKCCSZB15}, rely more on global image-text alignment capabilities. In contrast, other tasks, such as visual question answering (VQA)~\cite{DBLP:conf/iccv/AntolALMBZP15}, typically require the model to answer questions based on specific parts of the image, which necessitates not only global image-text alignment but also local image-text alignment capabilities. However, previous work has neglected these differentiated alignment requirements.

\end{itemize}

To effectively enhance the alignment capabilities, we propose a new multimodal large language model called AlignGPT. In the pre-training phase, we aim to make the model to understand different degrees of the image-text alignment relation. Specifically, instead of treating all image-text pairs equally, we divide image-text pairs into different groups according to the degrees of alignment of them and give an extra group label to each pair. This process is achieved by the help of CLIP scores~\cite{DBLP:conf/icml/RadfordKHRGASAM21}, where the higher scores indicate the higher degrees of alignment~\cite{DBLP:journals/corr/abs-2111-02114,DBLP:conf/icvgip/GroverMG22}. For example, in \cref{fig:CLIP_Similarity}, the degree of alignment of each image-text pair rises from left to right and the CLIP score of each pair also increases. Subsequently, we utilize these group labels as control signals to make the model to learn the representations of different alignment levels. During the instruction-tuning phase, the model is trained to dynamically combine these representations obtained by pre-training for the instructions of each task. In this process, we not only assign global alignment capabilities but also adaptively configure different local alignment capabilities according to the alignment needs for instructions of each task. The broad range of tests conducted demonstrates that our model achieves competitive performance across 12 benchmarks, as shown in ~\cref{fig:AlignGPT_example}.


Our contribution can be summarized as follows:

\begin{itemize}
\item We propose a new multi-modal large language model AlignGPT to elevate and empower the alignment capabilities of MLLMs.


\item We propose a novel alignment strategy that learns different alignment levels in the pre-training stage, and then adaptively combines these alignment levels in the instruction-tuning stage to meet the needs of alignment capabilities for different tasks.

\item We conduct evaluations across multiple academic benchmarks and multimodal instruction-following benchmarks. Extensive experimental results show that our proposed AlignGPT achieves competitive performance. Further analysis verifies the effectiveness of the model.
\end{itemize}


\section{Related Work}

In this section, we review the existing studies on large language models and visual language models.

\paragraph{Large Language Models.}

In the field of natural language processing, BERT~\cite{DBLP:conf/naacl/DevlinCLT19} and GPT-2~\cite{radford2019language}, as pioneering large pre-trained language models, marked a significant breakthrough in this technological direction. Their training on vast web text datasets demonstrated unprecedented language understanding and generation capabilities. Subsequently, the launch of GPT-3~\cite{DBLP:conf/nips/BrownMRSKDNSSAA20} further accelerated the development of this field, with its large model parameters and extensive training datasets showcasing exceptional abilities in few-shot learning, significantly enhancing task adaptability and flexibility. Following this, the introduction of InstructGPT and ChatGPT~\cite{DBLP:conf/nips/Ouyang0JAWMZASR22} focused on optimizing the efficiency and naturalness of interactions between models and humans, where InstructGPT enhanced the capability to execute precise instructions, and ChatGPT improved the conversational experience, making these models more fluent in human-computer communication. Meanwhile, as large language model (LLM) technology continued to evolve, emerging models like LLaMA~\cite{DBLP:journals/corr/abs-2302-13971} and GLM~\cite{du2022glm} began to make their mark. To equip these models with the ability to respond to human instructions similar to ChatGPT, research teams finetune LLaMA and GLM using high-quality instruction datasets, thereby further enhancing its capability to follow instructions, with representative projects such as Alpaca~\cite{taori2023alpaca}, Vicuna~\cite{chiang2023vicuna}, and ChatGLM~\cite{DBLP:conf/iclr/ZengLDWL0YXZXTM23}.

Although these models have made significant progress in interacting with humans through language, we recognize that human understanding and processing of complex information relies not only on language but also critically on visual and other sensory inputs. The observation has driven us to further explore more comprehensive visual-language models in order to more accurately simulate complex interactions between humans and the real world.

\paragraph{Visual Language Models.}

In recent years, multimodal large language models (MLLMs) have garnered increasing attention. The core of MLLMs lies in their ability to achieve cross-modal understanding and generalization. Most current models, such as LLaVA~\cite{DBLP:conf/nips/LiuLWL23a}, MiniGPT-4~\cite{DBLP:journals/corr/abs-2304-10592}, mPLUG-Owl~\cite{DBLP:journals/corr/abs-2304-14178}, Qwen-VL~\cite{DBLP:journals/corr/abs-2308-12966}, MiniGPT-v2~\cite{DBLP:journals/corr/abs-2310-09478}, NExT-GPT~\cite{DBLP:journals/corr/abs-2309-05519}, InternLM-XComposer~\cite{DBLP:journals/corr/abs-2309-15112}, CogVLM~\cite{DBLP:journals/corr/abs-2311-03079}, and MM1~\cite{DBLP:journals/corr/abs-2403-09611}, utilize a standard training framework consisting of two primary phases: pre-training and instruction-tuning. In the pre-training phase, the model utilizes image caption data to establish a rich understanding of cross-modal semantic knowledge. This training phase enables the model to comprehend and capture the correlation between images and text, establishing a solid foundation for subsequent stage. In the instruction-tuning phase, the model receives specific task instructions to optimize its performance on that task. Through this instruction-tuning phase, the model can further refine its understanding to execute specific tasks, enabling it to flexibly and accurately address various task requirements in practical applications.



Although current MLLMs have achieved promising results, they overlook two critical factors. First, the degree of alignment between different image-text pairs is inconsistent during the pre-training phase. Second, different tasks require different levels of alignment capabilities during the instruction-tuning phase. As a result, the modeling of alignment capabilities in these models remains inadequate. To address these limitations, we propose a new multimodal large language model AlignGPT to effectively enhance the alignment capabilities of MLLMs.
\begin{figure}[t]
\centering
    \includegraphics[width=0.45\textwidth]{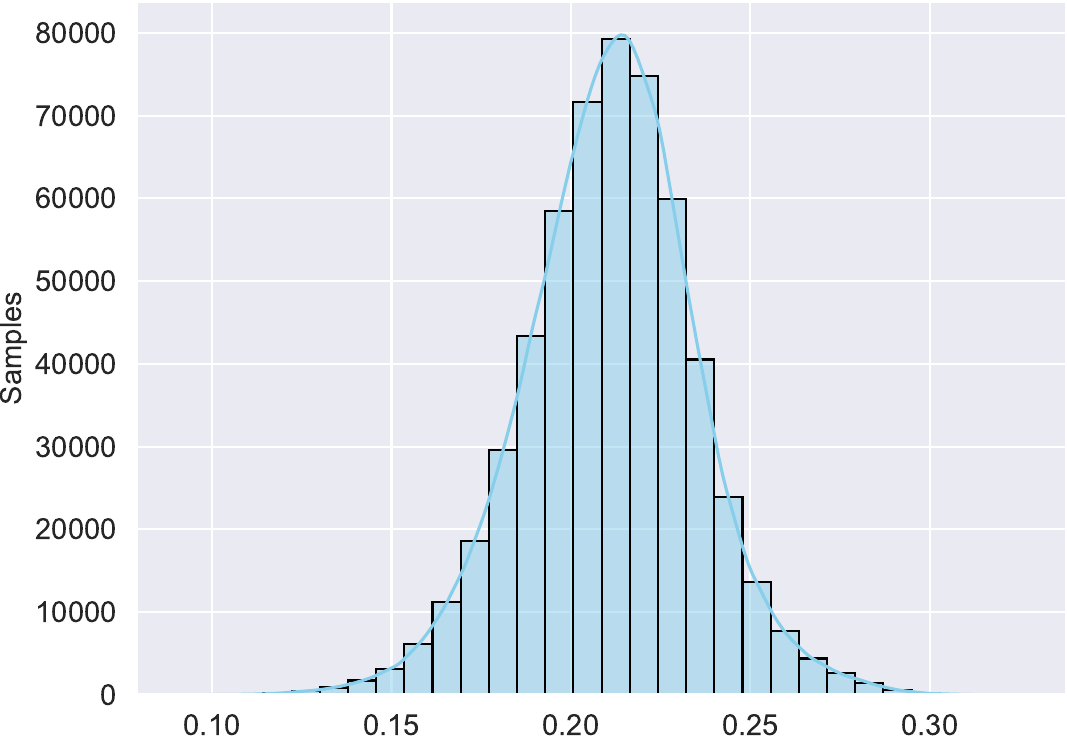}
  \caption{The distribution of CLIP similarity scores between images and texts in the pre-trained dataset.}
  \label{fig:CLIP_Similarity_dis}
\end{figure}

\section{Methodology}


In this section, we initially present the fundamental structure of the visual-language model AlignGPT, followed by a demonstration of how to enhance the alignment capability of the model by our pre-training and instruction-tuning paradigms.

\begin{figure*}[t]
\centering
    \includegraphics[width=0.88\textwidth]{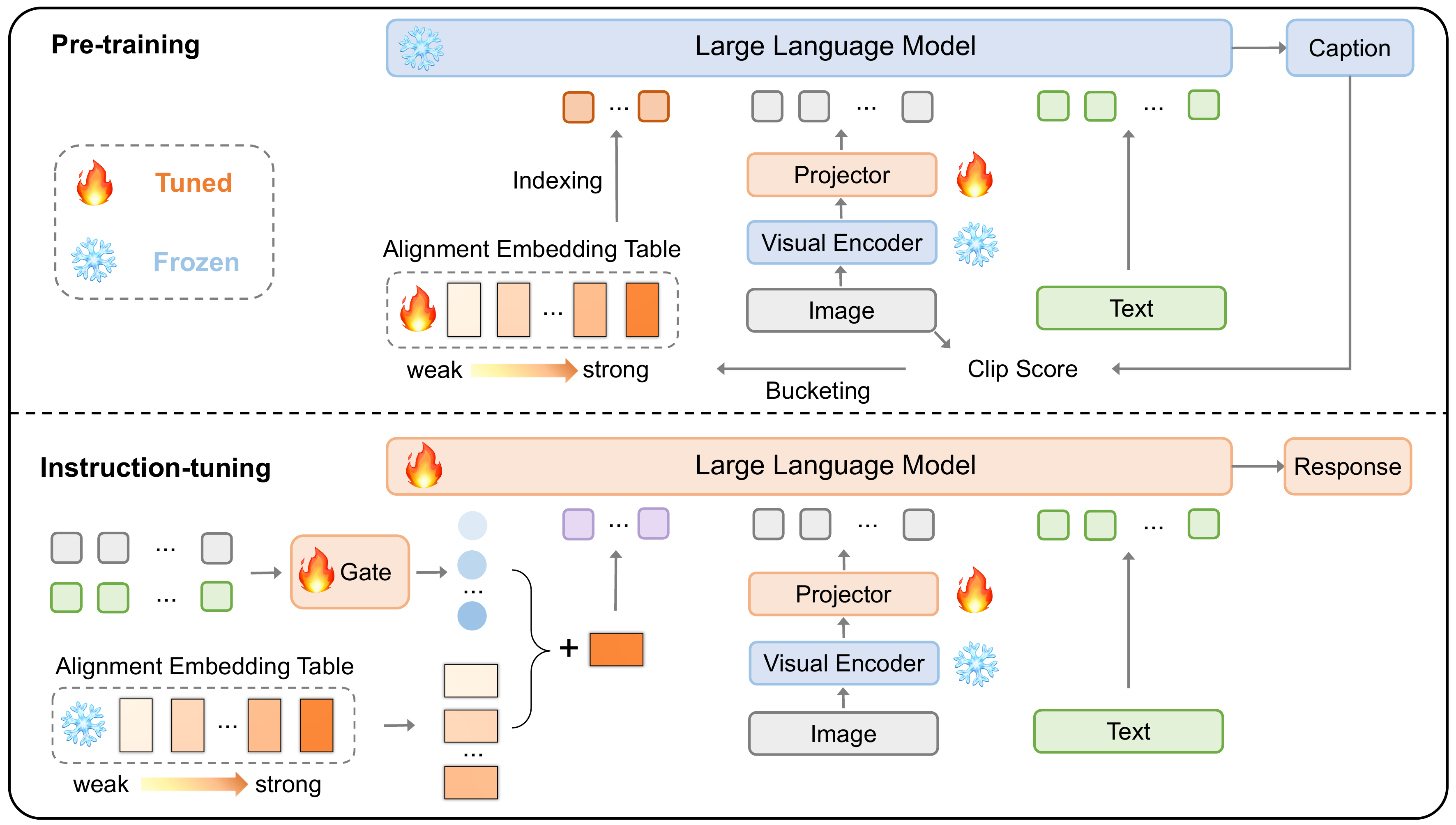}
  \caption{The architecture of AlignGPT.}
  \label{fig:AlignGPT_architecture}
\end{figure*}

\subsection{Architecture}

AlignGPT consists of four components: a visual backbone, a linear projection layer, a large language model, and an alignment module. ~\cref{fig:AlignGPT_architecture} provides an overview of the AlignGPT architecture and its training process. The followings are the implementation details of these components:

\paragraph{Visual backbone.}
We utilize the pre-trained CLIP visual encoder ViT-L/14~\cite{DBLP:conf/icml/RadfordKHRGASAM21} as our visual backbone. We train the model using an image resolution of 336$\times$336.

\paragraph{Linear projection layer.}
We adopt a linear projection layer to map the representations of images from the vector space of the vision backbone to that of the language model. 

\paragraph{Large language model.}
We choose the open-source model Vicuna~\cite{chiang2023vicuna} as our language model backbone, given its strong ability to follow instructions effectively in various language tasks.

\paragraph{Alignment module.}


We propose to add alignment embeddings to the inputs of MLLMs to enrich their alignment capabilities. These alignment embeddings are positioned ahead of the image embeddings and text embeddings. In the subsequent sections, we will elaborate on the role of the alignment embeddings and the process to acquire them.



\subsection{Alignment of Image and Text}

In our methodology, we utilize the similarity scores generated by the CLIP~\cite{DBLP:conf/icml/RadfordKHRGASAM21} model to evaluate the degree of alignment between images and text. As shown in \cref{fig:CLIP_Similarity}, we present four image-text pairs and the CLIP scores of them. From left to right, the degree of alignment between each image-text pair rises, i.e., the text description becomes more comprehensive. Correspondingly, the CLIP score of each pair increases. The rationale of adopting CLIP similarity scores lies in that CLIP is pre-trained on a massive dataset of paired images and their corresponding textual descriptions, which enables it to effectively capture the relationship between visual and linguistic information. By employing contrastive learning techniques~\cite{DBLP:conf/icml/ChenK0H20}, CLIP minimizes the distance between representations of similar image-text pairs while maximizing the distance between those of different pairs. This training approach relies on 400 million data pairs, allowing the model to develop a nuanced understanding of image-text relationships.



Beside, we also demonstrate the CLIP similarity distribution of image-text pairs in the pre-training dataset in \cref{fig:CLIP_Similarity_dis}. The results indicate that the CLIP similarity distribution varies significantly, suggesting a substantial difference in the alignment between images and texts. By jointly observing \cref{fig:CLIP_Similarity} and \cref{fig:CLIP_Similarity_dis}, we find that pairs with lower scores correspond to texts that describe only partial regions of the image, indicating weaker alignment. In contrast, pairs with higher scores reflect texts that provide a more comprehensive description of the image, suggesting a stronger alignment between the text and the image~\cite{DBLP:journals/corr/abs-2111-02114,DBLP:conf/icvgip/GroverMG22}.


\subsection{Alignment Level-aware Pre-training}\label{Pre-training}


As mentioned before, in the pre-training stage, the model usually assumes that all image-text pairs are uniformly aligned, and these pairs are used to train the model to comprehend the relations between images and their corresponding textual descriptions. However, in reality, the degree of alignment between these image-text pairs may vary considerably. Overlooking the difference could lead to a misunderstanding of the image-text alignment relations during the learning process.



Instead of treating all image-text pairs equally, we divide image-text pairs into different groups according to the degree of alignment of them and give each pair an extra group label. To achieve this, we leverage the similarity scores provided by CLIP. The higher the CLIP score is, the stronger the alignment is between image and text. Subsequently, we use these group labels as control signals to train the model, enabling it to understand the different alignment relations between different image-text pairs.


More precisely, we start by computing the CLIP similarities $s$ for all training image-text pairs. Subsequently, we rank all image-text pairs based on their similarity scores. Finally, we use a bucketing technique to divide them into $N$ discrete alignment levels. The process can be expressed as:
\begin{equation}
    l=bucket(s),\quad l \in \{1, 2, ..., N\},
\end{equation}
where $bucket(\cdot)$ denotes a bucketing function that assigns each pair into one of $N$ equally spaced intervals and $l$ is the alignment level (i.e., the group label) of an image-text pair. In this way, image-text pairs with lower CLIP similarity scores are assigned to buckets indicative of lower alignment levels, whereas those with higher CLIP similarity scores are grouped into buckets representing higher alignment levels. 

Once the alignment level of each image-text pair is determined, we can regard it as a special token to express the alignment relation between the image and its textual description. This special token is placed ahead of the image and text tokens. During the pre-training phase, in addition to learning the mapping function in the linear projection layer, we also initialize this special token as an alignment embedding and continuously update its representation.

\subsection{Adaptive Alignment-based Instruction-tuning}


Currently, the instructions used for finetuning cover various tasks such as image captioning, visual question answering, and visual grounding, etc. These tasks place different requirements on the alignment capabilities. For example, image captioning tasks mainly rely on global alignment between images and text, while VQA and visual grounding tasks require not only global alignment but also local alignment capabilities between images and text. To equip the model with the adaptive alignment capability, we propose an adaptive alignment-based instruction-tuning paradigm, which dynamically combine the alignment embeddings to meet the alignment needs for each task. 



To this end, we first clarify how to represent the global and local alignment capabilities between image-text pairs. As mentioned in Section \ref{Pre-training}, after the pre-training stage, we obtain $N$ alignment embeddings $\{H_1, H_2, ..., H_N\}$ corresponding to $N$ discrete alignment levels $\{1, 2, ..., N\}$. Among them, $H_N$ represents the highest level of alignment, \textit{i.e.}, $H_N$ indicates that the text provides very comprehensive description of an image. Here we regard it as a global alignment embedding. The embeddings below $H_N$ represent different degrees of alignment between the image and the text (\textit{i.e.}, $\{H_1, H_2, ..., H_{N-1}\}$), which means the text only describes a part of the information of the image from weak to strong. Thus, we regard them as local alignment embeddings of varying degrees.


Afterwards, we not only allocate global alignment capabilities to the instructions of each task, but also adaptively distribute varying degrees of local alignment capabilities based on the distinct alignment needs for each instruction. The reason behind this is that global alignment serves as the foundation for cross-modal understanding; only by mastering global alignment capabilities can a model truly focus on enhancing local alignment abilities. Specifically, in addition to the global alignment embeddings, we assign different weights to the local alignment embeddings via a gate network. These weights are obtained based on input instructions and image, as the input instructions greatly influence the visual regions the model should focus on. The implementation of the gate network is as follows:
\begin{equation}
    \alpha=softmax(W(H_I \otimes H_T)+b),
\end{equation}
where $H_I$ and $H_T$ denote the embeddings of the input instruction and the image, $W$ and $b$ are weight matrix and bias, $\alpha$ means the weights of local alignment embeddings. Finally, we aggregate the global alignment embedding and the local alignment embeddings with varying weights to ensure a more precise fulfillment of alignment requirements for instructions of each task:
\begin{equation}
    H_{align} = H_N + \sum_{i=1}^{N-1} \alpha H_i,
\end{equation}
where $H_{align}$ means the final alignment embedding for each instruction during the instruction-tuning stage.


In general, we can regard the alignment embeddings obtained in the pre-training phase as foundational components, each of which has different alignment capabilities. During the instruction-tuning phase, we dynamically combine these components to meet the alignment needs for instructions of different tasks.

\begin{table*}[t]
\renewcommand\arraystretch{1.2}
\centering
\scalebox{0.83}{
\begin{tabu}{l|l|c|ccccccc}
\toprule

{\textbf{Method}} & {\textbf{LLM}} & {\textbf{Resolution}} & {\textbf{POPE}} & {\textbf{MME}} & {\textbf{MMB}} & {\textbf{MMB$^{CN}$}} & {\textbf{SEED$^I$}} & {\textbf{LLaVA$^W$}} & {\textbf{MM-Vet}}\\

\midrule
BLIP-2 & Vicuna-13B & 224 & 85.3 & 1293.8 & - & - & 46.4 & 38.1 & 22.4 \\
InstructBLIP & Vicuna-7B & 224 & - & - & 36.0 & 23.7 &  53.4 & 60.9 & 26.2\\
InstructBLIP & Vicuna-13B & 224 & 78.9 & 1212.8 & - & - & - &  58.2 & 25.6\\
Shikra  & Vicuna-13B & 224 & - & - & 58.8 & - & - & - & -\\
IDEFICS-9B & LLaMA-7B & 224 & - & - &  48.2 & 25.2 & - & - & -\\
IDEFICS-80B & LLaMA-65B & 224 & - & - & 54.5 & 38.1 & - & - & -\\
MiniGPT-v2 & LLaMA2-7B & 448 & 85.1$^\diamondsuit$ & 1332.1$^\diamondsuit$ & 43.1$^\diamondsuit$ & 29.1$^\diamondsuit$ & 52.3$^\diamondsuit$ & - & -\\
Qwen-VL & Qwen-7B & 448 & - & - & 38.2 & 7.4 & 56.3 & - & -\\
Qwen-VL-Chat & Qwen-7B & 448 & - & 1487.5 & 60.6 &  56.7 &  58.2 & - & -\\
LLaVA-1.5 & Vicuna-7B & 336 & 85.9 & 1510.7 & 64.3 & 	58.3 & 	66.2 & 	63.4 & 30.5\\
LLaVA-1.5 & Vicuna-13B & 336 & 85.9 & 1531.3 & 67.7 & 63.6 & \textbf{68.2} & 70.7 & 35.4\\
\midrule
\rowcolor{gray!20} \textbf{AlignGPT} & Vicuna-7B & 336 & 86.0 & 1527.4 & 67.3 & 59.9 & 66.5 & 68.4 & 30.8\\ 
\rowcolor{gray!20} \textbf{AlignGPT} & Vicuna-13B & 336 & \textbf{86.2} & \textbf{1572.0} & \textbf{69.5} &  \textbf{63.7} & 67.8 & \textbf{75.2} & \textbf{35.6}\\
\bottomrule
\end{tabu}}
\caption{Results on multimodal instruction-following benchmarks. For the baseline methods, the results on the SEED$^I$ dataset are obtained from~\cite{DBLP:journals/corr/abs-2311-12793}, and the other results are retrieved from~\cite{DBLP:journals/corr/abs-2310-03744}.}
\label{main_result_instruction}%
\end{table*}

\begin{table*}[t]
\renewcommand\arraystretch{1.2}
\centering
\scalebox{0.83}{
\begin{tabu}{l|l|c|cc|ccccc}
\toprule
\multirow{2}{*}{\textbf{Method}} & \multirow{2}{*}{\textbf{LLM}} & \multirow{2}{*}{\textbf{Resolution}} & \multicolumn{2}{c|}{\textbf{Sample Size}} & \multirow{2}{*}{\textbf{VQA$^{v2}$}} & \multirow{2}{*}{\textbf{GQA}} & \multirow{2}{*}{\textbf{VisWiz}} & \multirow{2}{*}{\textbf{SQA$^I$}} & \multirow{2}{*}{\textbf{TextVQA}}\\
\multirow{-2}{*}{} & & & \textbf{Pre-train} & \textbf{Finetune}\\ 

\midrule
BLIP-2 & Vicuna-13B & 224 & 129M & - & 41.0 & 41.0 & 19.6 & 61.0 & 42.5\\
InstructBLIP & Vicuna-7B & 224 & 129M & 1.2M & - & 49.2 & 34.5 & 60.5 & 50.1\\
InstructBLIP  &  Vicuna-13B & 224 & 129M & 1.2M & - & 49.5 & 33.4 & 63.1 & 50.7\\
Shikra  & Vicuna-13B & 224 & 600K & 5.5M & 77.4 & - & - & - & -\\
IDEFICS-9B & LLaMA-7B & 224 & 353M & 1M & 50.9 & 38.4 & 35.5 & - & 25.9\\
IDEFICS-80B & LLaMA-65B & 224 & 353M & 1M & 60.0 &  45.2 &  36.0 & - &  30.9\\
MiniGPT-v2 & LLaMA2-7B & 448 & - & - & 74.6$^\diamondsuit$ & 60.3 & 32.9 & 60.9$^\diamondsuit$ & 28.0$^\diamondsuit$\\
Qwen-VL & Qwen-7B & 448 & 1.4B & 50M & 78.8 & 59.3 &  35.2 & 67.1 & \textbf{63.8}\\
Qwen-VL-Chat & Qwen-7B & 448 & 1.4B & 50M & 78.2 & 57.5 & 38.9 & 68.2 & 61.5\\
LLaVA-1.5 & Vicuna-7B & 336 & 558K & 665K & 78.5 & 62.0 & 50.0 & 66.8 & 58.2\\
LLaVA-1.5 & Vicuna-13B & 336 & 558K & 665K & \textbf{80.0} &  63.3 & 53.6 & \textbf{71.6} & 61.3\\
\midrule
\rowcolor{gray!20} \textbf{AlignGPT} & Vicuna-7B & 336 & 558K & 665K & 79.1 & 62.9 & 54.2 & 68.5 & 58.4\\ 
\rowcolor{gray!20} \textbf{AlignGPT} & Vicuna-13B & 336 & 558K & 665K & \textbf{80.0} & \textbf{63.6} & \textbf{56.4} & 70.3 & 60.2\\
\bottomrule
\end{tabu}}
\caption{Performance comparison on multiple academic benchmarks. For the baselines, the results with $^\diamondsuit$ are obtained by running the code released by the authors, and the other results are retrieved from~\cite{DBLP:journals/corr/abs-2310-09478,DBLP:journals/corr/abs-2310-03744}. Best results are in bold.}
\label{main_result_vqa}%
\end{table*}

\begin{table*}[t]
\renewcommand\arraystretch{1.2}
\centering
\scalebox{0.80}{
\begin{tabu}{l|c|ccccccccc}
\toprule
{\textbf{Method}} & {\textbf{Alignment Level}} & {\textbf{VQA$^{v2}$}} & {\textbf{GQA}} & {\textbf{VisWiz}} & {\textbf{SQA$^I$}} & {\textbf{TextVQA}} & {\textbf{POPE}} & {\textbf{MME}} & {\textbf{MMB}} &{\textbf{SEED$^I$}}\\


\midrule
AlignGPT & Number=4 & 79.0 & 62.9 & 52.3 & 68.7 & 	58.3 & 86.2 & 1463.8 & 67.2 & 66.5\\ 
AlignGPT & Number=6 & 79.0 & 62.7 & 51.2 & 68.9 & 	58.3 & 85.8 & 1436.3 & 67.3 & 66.2\\
AlignGPT & Number=8 & 79.1 & 62.9 & 54.2 & 68.5 & 58.4 & 86.0 & 1527.4 & 67.3 & 66.5\\
AlignGPT & Number=10 & 79.1 & 62.6 & 53.0 &	67.8 & 58.4 & 86.2 & 1481.4 & 66.4 & 66.7\\
\bottomrule
\end{tabu}}
\caption{Influence of different number of alignment level.}
\label{main_result_indicator}%
\end{table*}

\begin{table*}[t]
\renewcommand\arraystretch{1.2}
\centering
\scalebox{0.80}{
\begin{tabu}{c|ccc|cccccccccc}
\toprule
{\textbf{Settings}} & {\textbf{Average}} & {\textbf{Local}} & {\textbf{Global}} & {\textbf{VQA$^{v2}$}} & {\textbf{GQA}} & {\textbf{VisWiz}} & {\textbf{SQA$^I$}} & {\textbf{TextVQA}} & {\textbf{POPE}} & {\textbf{MME}} & {\textbf{MMB}} & {\textbf{SEED$^I$}}\\
\midrule
(a) & \xmark & \cmark & \xmark & 79.1 & 62.7 & 53.3 & 67.9 & 58.6 & 85.9 & 1467.1 & 66.9 & 66.3\\
(b) & \xmark & \xmark & \cmark & 79.1 & 62.9 & 52.6 &	68.3 & 58.4 & 85.9 & 1502.9 & 66.3 & 66.2\\
(c) & \cmark & \xmark  & \cmark & 79.0 & 62.8 & 52.5 & 68.6 & 58.4 & 85.6 & 1492.5 & 67.0 & 66.0\\ 
(d) & \xmark & \cmark & \cmark & 79.1 & 62.9 & 54.2 & 68.5 & 58.4 & 86.0 & 1527.4 & 67.3 & 66.5\\
\bottomrule
\end{tabu}}
\caption{Influence of local and global alignment.}
\label{main_result_components}%
\end{table*}

\section{Experiments}\label{Experiments}

\subsection{Experimental Settings}\label{Experimental_Settings}

\paragraph{Datasets.}

For a fair comparison, we use the same pre-training and instruction dataset as the LLaVA-1.5~\cite{DBLP:journals/corr/abs-2310-03744}. It mainly includes 558K caption pairs for modality alignment and 665K single- or multi-round conversations for instruction-tuning. Besides, we evaluate AlignGPT on a range of academic visual question answering (VQA) tasks and recent benchmarks designed specifically for MLLMs. This evaluation spans 12 benchmarks, including VQA$^{V2}$~\cite{DBLP:conf/cvpr/GoyalKSBP17}, GQA~\cite{DBLP:conf/cvpr/Gurari0SGLGLB18}, VizWiz~\cite{DBLP:conf/cvpr/Gurari0SGLGLB18}, SQA$^I$ (ScienceQA-IMG)~\cite{DBLP:conf/nips/LuMX0CZTCK22}, TextVQA~\cite{DBLP:conf/cvpr/SinghNSJCBPR19}, POPE~\cite{DBLP:conf/emnlp/LiDZWZW23}, MME~\cite{DBLP:journals/corr/abs-2306-13394}, MMB (MMBench), MMB$^{CN}$ (MMBench-Chinese)~\cite{DBLP:journals/corr/abs-2307-06281}, SEED$^I$ (SEED-Bench-IMG)~\cite{DBLP:journals/corr/abs-2307-16125}, LLaVA$^W$ (LLaVA-Bench-in-the-Wild)~\cite{DBLP:conf/nips/LiuLWL23a}, and MM-Vet~\cite{DBLP:journals/corr/abs-2308-02490} datasets.

\paragraph{Implementation Details.}

We adopt a ViT~\cite{dosovitskiy2021an} model pre-trained with CLIP~\cite{DBLP:conf/icml/RadfordKHRGASAM21} as a vision encoder to process visual inputs. On the language side, Vicuna~\cite{chiang2023vicuna} is utilized to handle multimodal features, ensuring a cohesive integration of text and visual data. In the pre-training phase, both the visual backbone and the large language model of AlignGPT remain frozen, with only the parameters of the linear projection layer and alignment embeddings being trained. During instruction-tuning phase, we freeze the alignment embeddings and visual backbone, while adjusting the parameters of the linear projection layer, large language model, and the gate network. The global batch sizes for the two phases are set at 256 and 128 respectively, with DeepSpeed~\cite{DBLP:conf/sc/RajbhandariRRH20} using ZeRO2 and ZeRO3 strategies accordingly. Regarding our training methodology, we conduct a single epoch of optimization for all models using the AdamW~\cite{loshchilov2018decoupled} optimizer coupled with a cosine learning schedule. Moreover, we initiate pre-training and instruction-tuning with learning rates of 1e-3 and 2e-5, respectively. The framework is trained on 8 A800 GPUs with 80GB memory.

\subsection{Compared Methods}


We chose a diverse set of representative MLLMs as our baselines, including BLIP-2~\cite{DBLP:conf/icml/0008LSH23}, InstructBLIP~\cite{DBLP:conf/nips/Dai0LTZW0FH23}, Shikra~\cite{DBLP:journals/corr/abs-2306-15195}, IDEFICS~\cite{DBLP:conf/nips/LaurenconSTBSLW23}, MiniGPT-v2~\cite{DBLP:journals/corr/abs-2310-09478}, Qwen-VL~\cite{DBLP:journals/corr/abs-2308-12966}, Qwen-VL-Chat~\cite{DBLP:journals/corr/abs-2308-12966}, and LLaVA-1.5~\cite{DBLP:journals/corr/abs-2310-03744}. 

\begin{table*}[t]
\renewcommand\arraystretch{1.2}
\centering
\scalebox{0.83}{
\begin{tabu}{l|l|c|ccccccc}
\toprule
{\textbf{Method}} & {\textbf{LLM}} & {\textbf{Resolution}} & {\textbf{VQA$^{v2}$}} & {\textbf{GQA}} & {\textbf{SQA$^I$}} & {\textbf{TextVQA}} & {\textbf{POPE}} & {\textbf{MMB}} &{\textbf{SEED$^I$}}\\

\midrule
AlignGPT & Vicuna-7B & 336 & 79.1 & 62.9 & 68.5 & 58.4 & 86.0 & 67.3 & 66.5\\ 
AlignGPT & Vicuna-7B & 672 & 79.7 & 63.3 & 68.3 & 60.3 & 86.8 & 67.2 & 66.5\\
AlignGPT & Vicuna-7B & 1008 & 79.8 & 63.4 & 68.2 & 60.3 & 86.8 & 67.2 & 66.6\\
\bottomrule
\end{tabu}}
\caption{Influence of different input image resolutions.}
\label{main_result_Res}%
\end{table*}

\begin{table*}[t]
\renewcommand\arraystretch{1.2}
\centering
\scalebox{0.83}{
\begin{tabu}{l|l|c|ccccccc}
\toprule
{\textbf{Method}} & {\textbf{LLM}} & {\textbf{Resolution}} & {\textbf{VQA$^{v2}$}} & {\textbf{GQA}} & {\textbf{SQA$^I$}} & {\textbf{MME}} & {\textbf{MMB}} & {\textbf{MMB$^{CN}$}} & {\textbf{SEED$^I$}}\\
\midrule
AlignGPT & LLaMA2-7B-Chat & 336 & 79.1 & 62.9 & 65.9 & 1500.8 & 66.6 & 57.9 & 66.4\\
AlignGPT & Vicuna-v1.5-7B & 336 & 79.1 & 62.9 & 68.5 & 1527.4 & 67.3 & 59.9 & 66.5\\ 
AlignGPT & LLaMA3-8B-Base & 336 & 79.6 & 63.1 & 70.4 & 1539.7 & 72.0 & 67.7 & 68.2\\
\bottomrule
\end{tabu}}
\caption{Influence of different large language models.}
\label{main_result_Language}%
\end{table*}

\section{Results and Analysis}

\subsection{Main Results}

\paragraph{MLLM-oriented Multi-modal Benchmarks.}

We apply AlignGPT to seven recent popular multimodal benchmarks, as shown in ~\cref{main_result_instruction}. We discover that, apart from LLaVA-1.5-13B, AlignGPT-7B surpassed all previous multimodal models. This shows that our model has strong generalization ability. Additionally, compared to LLaVA-1.5-13B, AlignGPT-13B has shown improvements on most datasets, particularly achieving good advancements on the MME, MMB, and LLaVA$^W$ benchamrks. This further validates the efficacy of both global and local alignment capabilities.

\paragraph{Visual Question Answering.}

We evaluate AlignGPT using five popular academic benchmarks, as detailed in ~\cref{main_result_vqa}. Despite using less training data, the AlignGPT-7B demonstrates competitive performance, surpassing other generalist models including InstructBLIP-13B, Shikra-13B, and IDEFICS-80B on most datasets, except for LLaVA-1.5-13B. These results verify the rationality of the structural design of our model. Moreover, considering that AlignGPT utilizes the same training dataset as LLaVA-1.5, it is evident that AlignGPT-7B outperforms LLaVA-1.5-7B across all evaluation datasets, and AlignGPT-13B also surpasses LLaVA-1.5-13B on the majority of datasets. This demonstrates that our approach effectively enhances the alignment capabilities of multimodal large language models. The fly in the ointment is that AlignGPT-13B does not perform as well as Qwen-VL on the TextVQA dataset. This may stem from the fact that TextVQA is a text-centric QA task, as it requires identifying text in images to answer questions. AlignGPT is tailored to boost multimodal alignment and might not exhibit strong results in text-focused scenarios.

\begin{figure}[t]
\centering
     \includegraphics[width=0.46\textwidth]{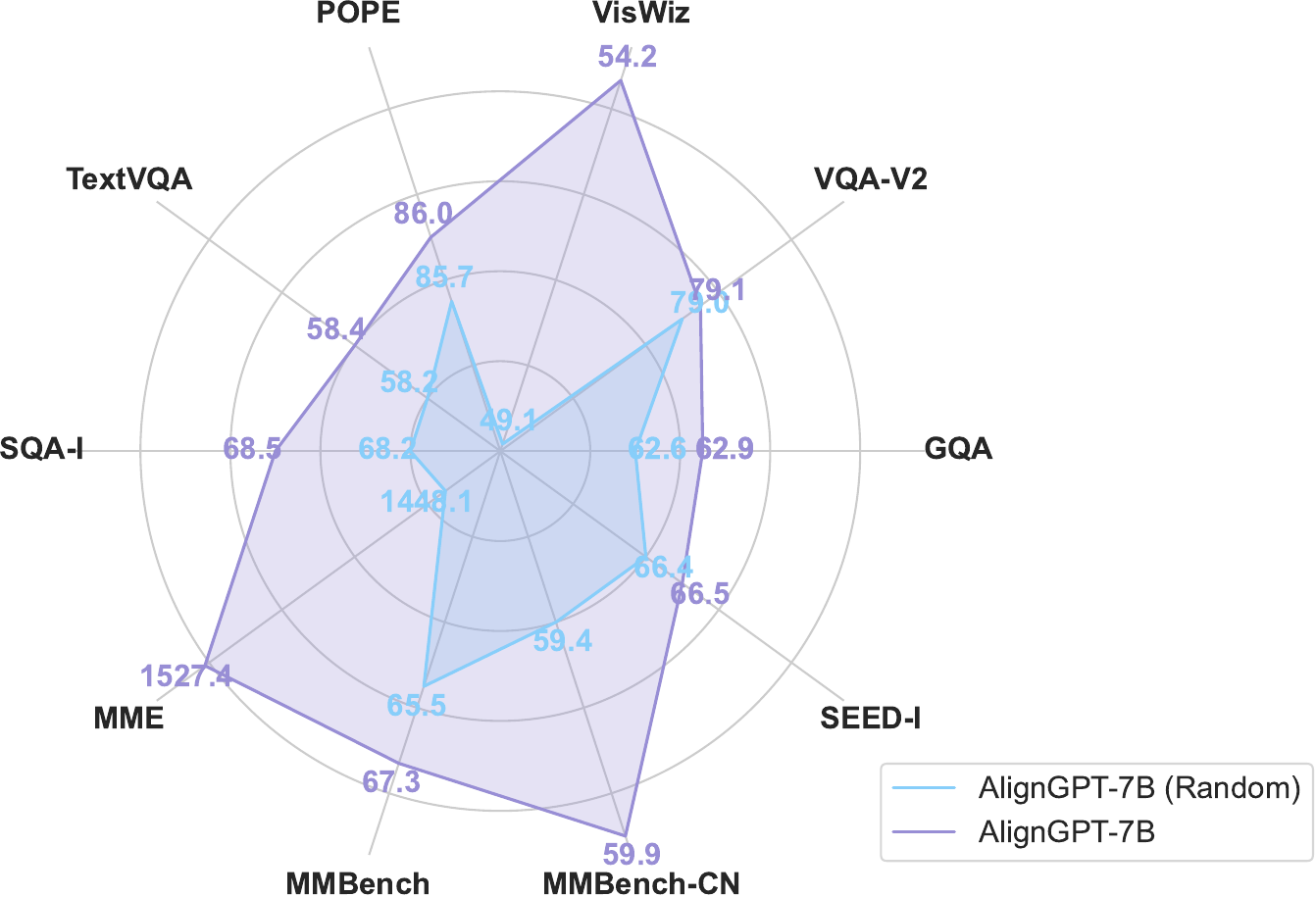}
  \caption{The performance comparison of AlignGPT (random) and AlignGPT in downstream datasets.}
  \label{fig:Alignment_table}
\end{figure}

\begin{figure*}[t]
    \centering
    \begin{subfigure}[t]{0.30\textwidth}
        \centering
        \includegraphics[width=\textwidth]{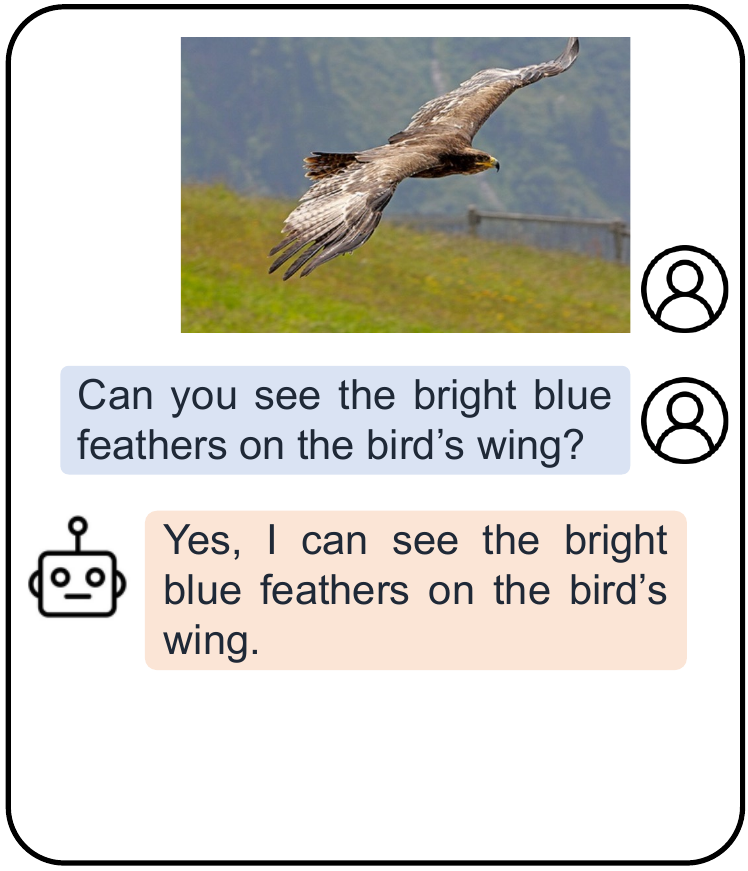}
        \caption{MiniGPT-v2}
    \end{subfigure}
    \begin{subfigure}[t]{0.30\textwidth}
        \centering
        \includegraphics[width=\textwidth]{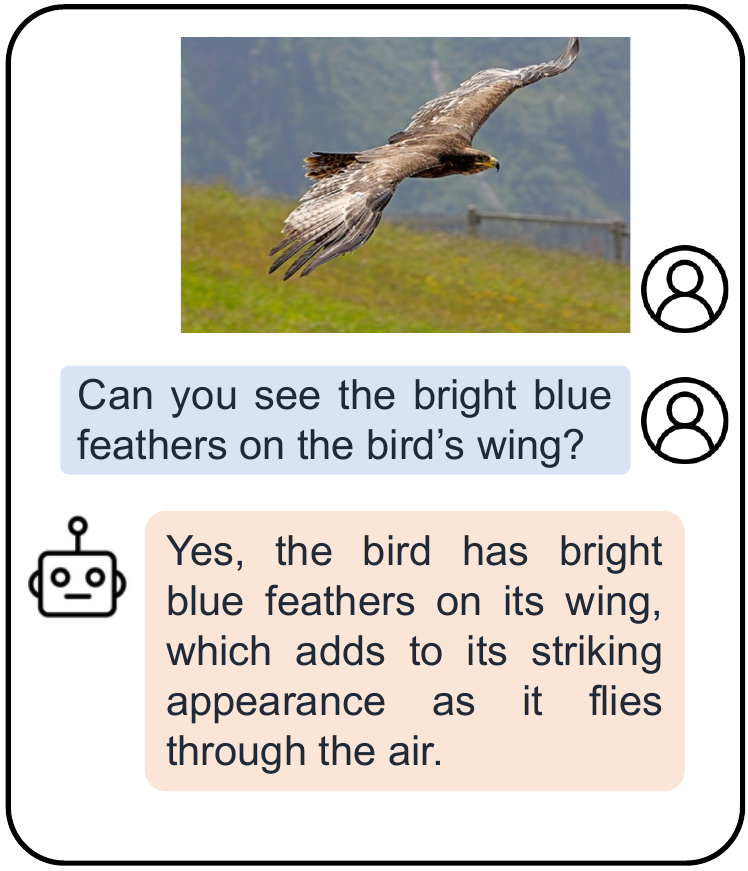}
        \caption{LLaVA-1.5}
    \end{subfigure}
    \begin{subfigure}[t]{0.30\textwidth}
        \centering
        \includegraphics[width=\textwidth]{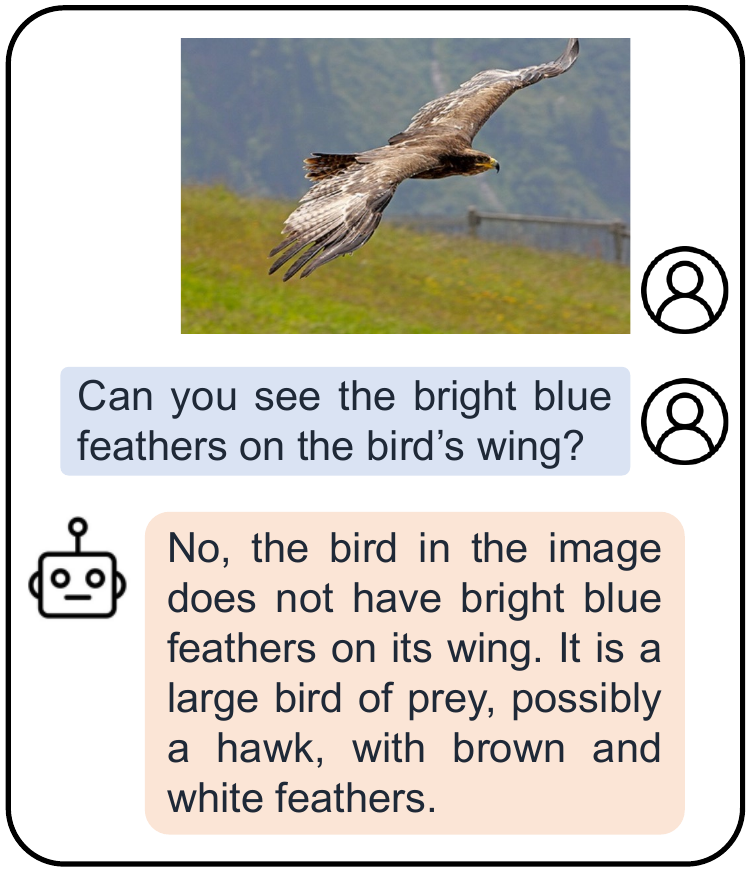}
        \caption{AlignGPT}
    \end{subfigure}
    \caption{Comparison with other models.}
    \label{fig:case_example}
\end{figure*}

\begin{figure*}[t]
    \centering
    \begin{subfigure}[t]{0.30\textwidth}
        \centering
        \includegraphics[width=\textwidth]{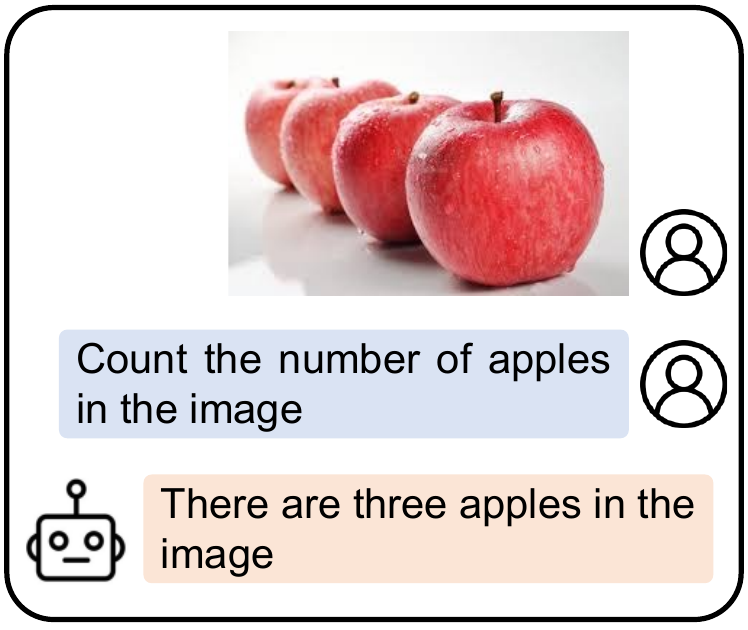}
        \caption{alignment level=1}
    \end{subfigure}
    \begin{subfigure}[t]{0.30\textwidth}
        \centering
        \includegraphics[width=\textwidth]{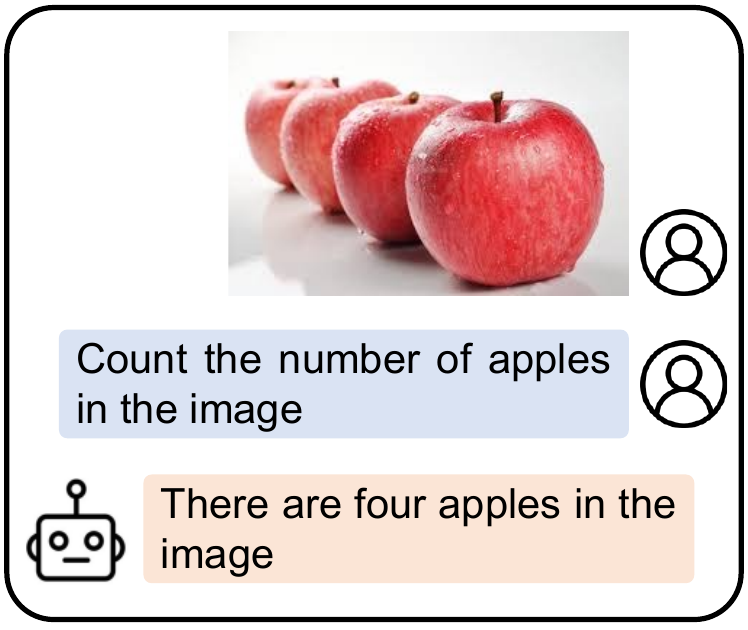}
        \caption{alignment level=4}
    \end{subfigure}
    \begin{subfigure}[t]{0.30\textwidth}
        \centering
        \includegraphics[width=\textwidth]{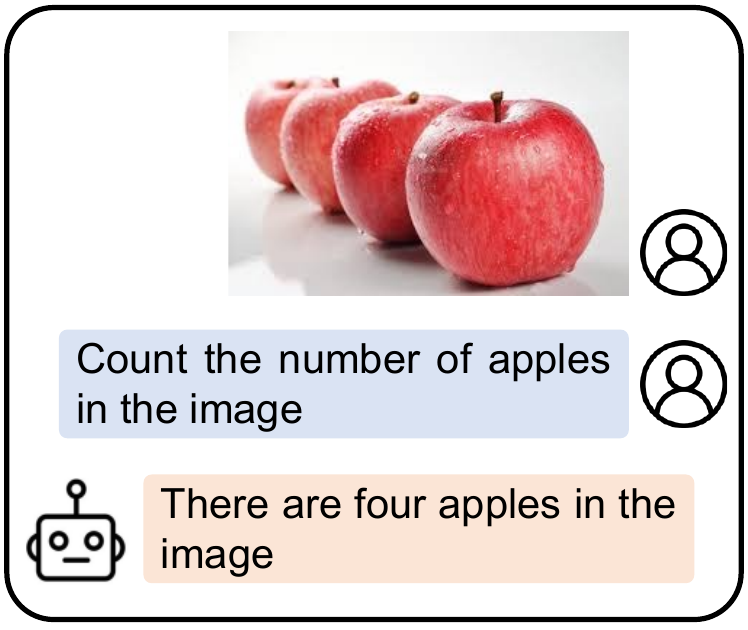}
        \caption{alignment level=7}
    \end{subfigure}
    \caption{The responses of AlignGPT under different levels of alignment capability.}
  \label{fig:different_alignment}
\end{figure*}

\subsection{Ablation Study}

Without loss of generality, we choose AlignGPT-7b for the ablation study to analyze the impact of various components.


\paragraph{Impact of Alignment Embedding Table.}

We design an experiment to validate the role of the alignment embedding table. In the fine-tuning phase, we use a randomly initialized alignment embedding table instead of the embedding table obtained during the pre-training phase. Since both methods have the same number of parameters, we can clearly assess whether the parameter number of the alignment embedding table influences the improvement in model performance. The experimental results are shown in ~\cref{fig:Alignment_table}. We find that the model using the randomly initialized alignment embedding table (referred to as AlignGPT (random)) shows a performance gap compared to AlignGPT. These results further confirm that the alignment information learned during pre-training is indeed the key factor in enhancing model performance, rather than the parameter number.

\paragraph{Impact of Number of Alignment Levels.}

To investigate the effect of the number of alignment levels $N$ on AlignGPT, we vary the value of $N$ in the range of [4, 10] with a step size of 2.  ~\cref{main_result_indicator} shows the performance of AlignGPT with different $N$ on nine datasets. Actually, AlignGPT can achieve good results at $N=4$, and their performance remains stable as the number of alignment levels increases. Depending on the trajectory of the curve, their performance has an initial upward trend and then flattens out. These observations indicate that AlignGPT can improve the alignment capabilities of multi-modal large language models based on a small number of alignment levels. Finally, according to the trend of the curve, we set $N$ to 8.

\paragraph{Impact of Local and Global Alignment.}


During the instruction-tuning phase, we assign global and local alignment capabilities to the instructions of each task. Among them, ``Local'' refers to the local alignment capabilities derived by assigning different weights to various local alignment embeddings using a gate network. ``Global'' denotes the global alignment capabilities, and ``Average'' represents the local alignment capabilities obtained by assigning equal weights to each local alignment embedding. The performance of these four strategies (settings a-d) is presented in ~\cref{main_result_components}. As we can see, setting (a) and setting (b) demonstrate divergent performances in downstream tasks, which can be attributed to the different demands these tasks place on global and local alignment capabilities. It is worth noting that setting (a) and setting (b) perform worse than our final approach (setting d) on most datasets, which verifies the necessity of the combination of global and local alignment capabilities. Moreover, the performance of setting (c) is inferior to that of setting (d), which may be due to the dynamically changing demands for local alignment capabilities across different downstream tasks.

\subsection{Discussion}

\paragraph{Impact of different input image resolutions.}


Image resolution plays a crucial role in vision-language tasks as higher resolutions help reduce image blurring and enhance the understanding of image-text alignment. To evaluate the impact of resolution changes on the performance of multimodal tasks, we increase the image resolution from 336 to 1008, with the resulting performance changes detailed in ~\cref{main_result_Res}. The study results show that higher image resolutions can improve model performance on most multimodal tasks. For example, the score for VQA$^{v2}$ increased from 79.1 to 79.8, while the score for TextVQA rose from 58.4 to 60.3. Meanwhile, the performance of the POPE improve by 0.8. These results highlight that appropriately increasing image resolution is an effective strategy for enhancing performance in studies of multimodal large language models.

\paragraph{Impact of different large language models.}

We also explore the impact of the large language model on the performance of AlignGPT, specifically testing three models: LLaMA-2-7B-Chat, Vicuna-v1.5-7B, and the latest LLaMA-3-8B-Base. The results are shown in ~\cref{main_result_Language}. Initially, we observe that LLaMA-3-8B-Base achieves the best performance, followed by Vicuna-v1.5-7B, with LLaMA-2-7B-Chat performing the worst, which is reasonable given LLaMA-3-8B-Base's larger parameter size and richer training data. Besides, we observe that Vicuna-v1.5-7B achieves superior performance over LLaMA-2-7B-Chat on multimodal benchmarks such as MME, MMB, and SEED$^I$, while showing comparable results on VQA tasks. This advantage might be due to Vicuna-v1.5-7B undergoing supervised instruction-tuning with ShareGPT data, which contains background knowledge relevant to downstream tasks.

\subsection{Qualitative Results}

Figure~\ref{fig:case_example} presents a comparative analysis of our model with MiniGPT-v2~\cite{DBLP:journals/corr/abs-2310-09478} and LLaVA-1.5~\cite{DBLP:journals/corr/abs-2310-03744}. When a user submits an image alongside the instruction ``Can you see the bright blue feathers on the bird's wing?'', MiniGPT-v2 and LLaVA-1.5 both return an incorrect answer ``Yes''. In contrast, our model produces accurate result ``No'', thereby demonstrating that AlignGPT can effectively enhance the model's alignment capability. In Figure~\ref{fig:different_alignment}, we further demonstrate the responses of AlignGPT under different levels of alignment capability. We find that with lower alignment levels, the model may only focus on certain regions of the image, resulting in an undercount of the total number of apples; whereas with higher alignment levels, the model considers the entire image area, thus achieving accurate apple quantity estimation. This finding once again underscores the necessity of enhancing the alignment capability of MLLMs.

\section{Conclusion}

In this paper, we propose AlignGPT, a novel multimodal large language model designed to bolster the alignment capabilities of MLLMs. Our approach involves utilizing the alignment level of data as a control signal during pre-training to effectively handle the varying degrees of alignment in image-text pairs. Subsequently, in the instruction-tuning phase, we begin by exploiting these control signals to shape different levels of alignment capabilities. Continuing from this, we go beyond assigning global alignment capabilities to instructions of each task; we also dynamically configure distinct local alignment capabilities based on the specific demands of each instruction. Results from numerous experiments indicate that our AlignGPT achieves better performance than other state-of-the-art MLLMs. 


\section*{Limitations}
The current study has two limitations: (1) This paper involves two modalities, i.e., text and image, while achieving AGI should also encompass video and audio, which requires us to do further research and exploration; (2) We propose a new perspective to enhance the alignment capability of MLLMs. However, there may be other methods to achieve this goal, which merit consideration in the future.

{
    \small
    \bibliographystyle{ieeenat_fullname}
    \bibliography{main}
}


\end{document}